\documentclass[9pt,twocolumn]{article}

\usepackage[square,numbers]{natbib}
\usepackage[
    head=13pt, 
    foot=2pc,
    paperwidth=8.5in, 
    paperheight=11in,
    columnsep=24pt,
    top=52pt, 
    bottom=75pt, 
    inner=52pt, 
    outer=52pt,
    marginparwidth=2pc,
    heightrounded
]{geometry}

\usepackage{amsmath}
\usepackage{amssymb}
\usepackage{amsthm}
\newtheorem{definition}{Definition}

\usepackage{authblk}
\usepackage{hyperref}
\usepackage{subcaption}
\usepackage{graphicx}

\newcommand{\email}[1]{\texttt{\href{#1}{#1}}}

\begin{document}

\title{Avoiding Resentment Via Monotonic Fairness}
\date{}

\author[1]{Guy W. Cole}
\author[2]{Sinead A. Williamson}
\affil[1]{\email{guywcole@utexas.edu}}
\affil[2]{\email{Sinead.Williamson@mccombs.utexas.edu}}
\affil[1,2]{Department of Statistics and Data Science, The University of Texas at Austin, 2317 Speedway G2500, Austin, TX 78712}

\maketitle

\begin{abstract}
Classifiers that achieve demographic balance by explicitly using protected attributes such as race or gender are often politically or culturally controversial due to their lack of individual fairness, i.e.\ individuals with similar qualifications will receive different outcomes. Individually and group fair decision criteria can produce counter-intuitive results, e.g.\ that the optimal constrained boundary may reject intuitively better candidates due to demographic imbalance in similar candidates. Both approaches can be seen as introducing individual resentment, where some individuals would have received a better outcome if they either belonged to a different demographic class and had the same qualifications, or if they remained in the same class but had objectively worse qualifications (e.g.\ lower test scores). We show that both forms of resentment can be avoided by using monotonically constrained machine learning models to create individually fair, demographically balanced classifiers.
\end{abstract}

\section{Introduction}

Machine learning algorithms trained to infer relationships, classify individuals or predict individuals' future performance tend to replicate biases inherent in the data \citep{Caliskan:Bryson:Narayanan:2017,Bornstein:2018,Angwin:Larson:Muttu:Kirchner:2016}. Worse, when these algorithms are used as tools in policy decision making, they can form parts of feedback loops that magnify discriminatory effects. For example, predictive policing algorithms aim to predict where crimes will take place, but are trained on data from where crimes are reported or arrests are made -- which can be skewed by biased policing and might not reflect the true crime map. If police officers are sent to areas with high predictive crime rate, they will tend to make more arrests there, increasing the algorithm's confidence and amplifying discrepancies between the crime rate and the arrest rate \citep{Ensign:2018,Lum:Isaac:2016}. 

This tendency can be counteracted by designing algorithms that aim to yield similar accuracy across different demographics. One approach is to design algorithms that explicitly use information about the protected variable in developing the algorithm, whether by transforming the attributes of each demographic group \citep{dwork2012fairness}, learning embeddings that transform each demographic group to comparable representations \citep{MadCrePitZem2018,ZemQiSwePitDwo2013}, or training separate classifiers on each group \citep{DwoImmKalTauLei2018}. 

While these approaches are powerful tools for combating systemic inequalities, algorithms that aim for demographic fairness can appear unfair or opaque on the individual level. For example, we can achieve demographic fairness in college admissions by applying different cutoffs for different groups, but individuals below the cutoff for their demographic group but above the cutoff for a different demographic group will feel unfairly treated. Even if the different cutoffs can be justified on a population level---for example, if certain demographic groups have statistically disparate access to educational resources, leading to lower average test scores---they are often unpopular among the class with the stricter cutoffs, and can result in complaints and legal action. For example, Universities' affirmative action policies have frequently been the target of legal action from students who feel that they have been unfairly denied entry when compared with similarly qualified members of other ethnic groups, both past \citep{2013fisher,2016fisher,1978Bakke} and ongoing \citep{2014SFFA}. In practice, this often means that we must pick a single decision boundary for all groups, even if this limits the fairness of the resulting outcome.

Conversely, algorithms that exhibit individual fairness---where two similar individuals are treated similarly even if their demographic group differs---can easily propagate unfairness on a population level. Schools are often highly racially segregated due to location, and schools in wealthy, majority-white neighborhoods tend to have more resources and funding, which are in turn correlated with better academic performance in high school \citep{USCivilRights:2018,national2013nation}. This better performance in high school does not necessarily translate to better performance at the university level \citep{VidalRodeiro:Zanini:2015}.

Further, even within an individually fair system, individuals might still feel resentment towards their peers. Individual fairness can be seen as minimizing resentment between two individuals with similar attributes but different demographic group memberships: neither individual feels they would have had a more favorable outcome if they could switch their membership. However, it can still lead to resentment between two individuals with different attributes, if those attributes admit a natural ordering: if student A has a higher SAT score than student B and is identical on all other axes, student A would feel resentment if student B had the higher acceptance probability. This can amplify demographic discrepancies if the demographic-specific attribute distributions differ: if the SAT scores of a minority group trended notably higher than SAT scores of a majority group, an admissions system could still satisfy individual fairness while accepting primarily low-scoring individuals.

The goal of this paper is to automatically design decision rules that avoid individual resentment---both resentment towards someone with similar attributes but a different demographic group membership, and resentment towards someone with ``worse'' attribute values---while minimizing population-level unfairness. We demonstrate that this approach allows us to design rules that trade off predictive accuracy with group notions of fairness, while avoiding perceived unfairness on an individual level.

\section{Notions of fairness}\label{sec:bg}
We consider models for individuals characterized by some set of protected or sensitive attributes $A_i \in \mathcal{A}$ and non-protected attributes $X_i \in \mathcal{X}$. Our goal is to predict some outcome $Y_i$; in this paper we focus on binary classification problems where $Y_i \in \{0, 1\}$, but our approach can easily be applied in a regression setting where $Y_i\in \mathbb{R}$.

Protected attributes might be race or gender; we assume in this paper that these attributes are categorical, but this assumption can be relaxed. Non-protected attributes include other information relevant to decision making, such as test scores or credit history. These attributes might be highly correlated with our protected variables (for example, attending a historically black university is highly correlated with race), meaning that we cannot avoid unfair outcomes simply by excluding the protected attributes from our analysis (sometimes referred to as fairness through unawareness \citep{dwork2012fairness}).

        Definitions of fairness in machine learning are generally (but not exclusively) divided into two camps based on their level of attention: group-level fairness and individual-level fairness.
        
        \textbf{Individual fairness} aims to ensure that two individuals $u$ and $v$ with non-protected attributes $X_u, X_v$ have similar outcomes if $X_u$ and $X_v$ are similar, even if their protected attributes differ. Concretely, \citep{dwork2012fairness} describes a score function $f$ as individually fair if it is Lipschitz-continuous w.r.t.\ some metric $\mathcal{D}$ on $X$, i.e.\
        \begin{equation}d(f(X_u), f(X_v)) \le \mathcal{D}(X_u, X_v) ~\forall~ u, v \in \mathcal{U}\label{eqn:IndFair}\end{equation}
        where $d$ is a metric on the space of outcomes. This encapsulates the notion that if two individuals are similar in terms of non-protected attributes, they should have similar outcomes. We can think of individual fairness as avoiding \textit{resentment} w.r.t.\ the protected variable: Under an individually fair algorithm, no-one would have achieved a better solution if they had a different protected variable.

        Conversely, \textbf{group fairness} metrics aim to minimize population-level imbalances. For example, the notion of demographic parity \citep{dwork2012fairness} requires that the predicted outcome $\hat{Y}$ is independent of the protected variable $A$. Equalized odds \citep{HarPriSre2016} requires that the predicted outcome $\hat{Y}$ is independent of $A$ conditioned on the true outcome $Y$, allowing our predictor to depend on $A$ via $Y$. Equalized opportunity \citep{HarPriSre2016} relaxes this condition in a classification task where the outcome $\hat{Y}=1$ is seen as more desirable than $\hat{Y}=0$, to require conditional independence between predictor $\hat{Y}$ and protected variable $\hat{A}$ only when $Y=1$. Agarwal et al. \cite{AgaBeyDudLanWal2018} show that demographic parity, equalized odds, and their variants can be expressed in terms of a set of linear constraints.  In many cases, individual notions of fairness are at odds with group notions of fairness. For example, \cite{dwork2012fairness} shows that individually fair functions achieve perfect demographic parity if and only if the distribution over individuals is similar across demographic groups.
        
        A number of approaches attempt to balance individual and group notions of fairness. Dwork et al. \cite{dwork2012fairness} combine demographic parity with a relaxed notion of statistical parity, where members of group $A'$ are first mapped to match the distribution of group $A$ via a Lipschitz-continuous mapping. Later work expands this idea by mapping individuals' protected and non-protected attributes into some latent embedding or representation that is uninformative of the protected attribute \citep{ZemQiSwePitDwo2013,MadCrePitZem2018}. Using such a mapping can lead to individual resentment w.r.t.\ the protected attribute, however,  since changing an individual's protected attribute value would change its embedding, and hence its outcome.
        
        An alternative approach is to learn a single classifier on $X$ to predict $Y$, and to encourage fairness by regularization using a fairness-promoting penalty \citep{KamAkaSak2011,KamAkHidSak2012,berk2017convex} or constraints \citep{zafar2017parity,zafar2017aistats,AgaBeyDudLanWal2018}. If the classifiers used are Lipschitz-continuous, then they are all individually fair, since each individual is subject to the same classification function. The form of this function is governed by a trade-off between predictive accuracy, and some appropriate measure of (group-level) fairness. While this trade-off means regularization approaches may achieve lower accuracy and/or group-level fairness than representation-based approaches, their individual fairness yields transparency in implementation and avoids situations where individuals would have different outcomes under counterfactual protected attributes.

        Our approach builds upon this family of regularization-based algorithms. We introduce a new measure of fairness that protects against counterfactual resentment w.r.t.\ shifts in both protected and non-protected variables, even outside the training set. Loosely, our idea of monotonic fairness protects against two sources of resentment: the perception that one would have been better off in a different demographic group, and the perception that one would have been better off had they under-performed along a given axis.

        Our work also complements a body of work which explores definitions of fairness in which groups are collectively satisfied \cite{zafar2017parity,heidari2018}, with variations on being \textit{a priori} ambivalent or being \textit{a posteriori} free of desire to switch labels as a group.  These variations deal with the idea of resentment at a class level, while we examine it at an individual level.  
        
        Others have considered the idea of individual-level comparisons; Balcan et al.\ \cite{balcan2018} explore the concept of "envy freeness" in classification in the context of individual-specific utility functions, where a classifier can be optimal when no individual's utility function would be higher if they received the predicted outcome (or distribution of outcomes) given to an individual with different attributes. This approach could not be applied to settings where the utility function is assumed to be identical among individuals, e.g. in most classification tasks where this is a preferred outcome that all individuals would prefer.
        
        ``Meritocratic fairness"~\cite{joseph2016} appears similar, but differs in that it ranks points based on the expected outcome for their attribute values rather than the actual attribute, i.e.\ it is monotonic w.r.t.\ the expected true outcome rather than the predictors so that (in one form) if $\mathbb{E}[Y | X_u] > \mathbb{E}[Y | X_v]$ then $\hat{f}(X_u) \ge \hat{f}(X_v)$).  Our approach differs in that we require monotonicity w.r.t.~those inputs believed to directly correlate with performance (detailed in section \ref{sec:defs}).
        
        Lipton et al.~\cite{lipton2018treatmentdisparity} study concepts of impact disparity and treatment disparity which overlap our own. Their concept of impact disparity is similar to statistical parity, that protected classes should be treated similarly overall.  They conceive of treatment disparity similarly to our own {\it class resentment}, that individuals' treatments differ based on their protected class.  Our work expands on this to incorporate {\it score resentment}, and proposes and evaluates a concrete framework for structurally enforcing protection.
        
        Others have considered the problem of monotonicity in fair methods.  \citep{kearns2017meritocratic} explores the notion of monotonicity in the context of combining rankings between groups which lack common attributes, e.g.\ when comparing the athleticism of athletes from different sports.  Their method assumes that a perfect ranking is known within each sport, and compares athletes across sports using the sport-specific CDF of the outcome variables. Our method does not assume such a CDF estimate is obvious or accessible, and will not produce a separate classifier for each class of examples.  Similarly, \citep{dwork2018decoupled} consider decoupled classifiers for separate classes, and how they can be combined to produce fair classification.  Our model does not learn separate classifiers, which can introduce resentment between classes, but instead seeks to learn a unified classifier which satisfies fairness and prediction goals.

\section{Monotonic fairness}\label{sec:defs}

    Consider a model that outputs a score $f(X, A)$ to an individual with non-protected attributes $X$ and protected attribute $A$, where higher scores in some dimensions of $X$ are seen as more desirable.
    An example of $X_u$ being ``better'' than $X_v$ might be if the non-protected attributes correspond to SAT score, with $X_u$ being the higher score. 
    
    We assume in the remainder of this paper that non-protected attributes $X$ can be represented in $\mathbb{R}^d$. In general, we can subdivide $X$ into $X^+$ and $X^\circ$, where $X^+$ contains variables like SAT score, where certain values are deemed better than others, and $X^\circ$ variables like number of years in current position, where we do not wish to impose such value judgements. 

    Our paper considers the concept of \textit{individual resentment}, which can take the form of either {\it class resentment} and/or {\it score resentment}, which we define below. 
    
    \begin{definition} \label{def:ClassResentment}
        {\bf Protected Attribute Resentment (Class) Resentment:} Individual $u$ experiences {\it class resentment} under function $f$ if $\exists~A'$ s.t. $f(X_u, A_u) < f(X_u, A')$.
    \end{definition}
    Class resentment occurs when an individual who differs from another only in protected attributes receives a less-preferred outcome than that other individual, despite having identical non-protected attributes.  Even though there may be justifiable reasons for the discrepancy, the first individual is likely to perceive the system as penalizing them for their protected attribute.

    \begin{definition} \label{def:ScoreResentment}
        {\bf Non-Protected Attribute (Score) Resentment:} Individual $u$ experiences {\it score resentment} under function $f$ if there exists $(X', A')$ such that $X_u$ is objectively ``better" than $X'$ but $f(X_u, A_u) < f(X', A')$.
    \end{definition}
    
    Score resentment captures the situation where an individual receives a less-preferred outcome than another individual who differs only in having ``worse" scores in some dimensions -- for example, a candidate being rejected for being over-qualified for a job. While score resentment is typically not encoded into hand-designed systems, it can easily appear in automatically learned systems, as we discuss later in this section.

    Individually fair methods ensure that two individuals with similar non-protected attributes receive similar outcomes, avoiding the situation where an individual feels he or she would have been better treated had they belonged to a different demographic group---what we refer to above as protected attribute, or class, resentment. However, individual fairness does not necessarily avoid \textit{non-protected} attribute, or score, resentment--- the situation where an individual feels he or she would have been better treated had they performed worse on some axis. 
    
    We can ensure a score function has zero individual resentment by requiring that the function does not take the protected attribute as an input (guaranteeing zero protected attribute resentment) and is monotone non-decreasing w.r.t.\ all non-protected attributes in $X^+$ (guaranteeing zero non-protected attribute resentment). We refer to such a score function as being \textit{monotonically fair}.
    
        \begin{definition}
                {\bf Monotonic Fairness}: A function $f: \mathcal{X} \times \mathcal{A} \rightarrow \mathbb{R}$ is {\it monotonically fair} if no possible individual $(X, A) \in \mathcal{X}\times \mathcal{A}$ experiences {\it class resentment} (Def~\ref{def:ClassResentment}) or {\it score resentment} (Def~\ref{def:ScoreResentment}).
            \end{definition}

    To understand the difference between individual fairness and monotonic fairness, consider a system that admits students to college on the basis of a single standardized test. If the predictor is not non-decreasing w.r.t.\ that test result, a student could be in the unfair situation where they would have been accepted if their test result were lower. Similarly, a loan applicant might find themselves rejected for borrowing \textit{less} money. Such a predictor could arise, even if the true relationship between test score and probability of college success is monotonic, if our training data is sparse or demographically imbalanced in some area of the attribute space and especially in higher dimensional settings.
    
    \begin{figure*}[ht]
        \centering
        \includegraphics[width=\textwidth]{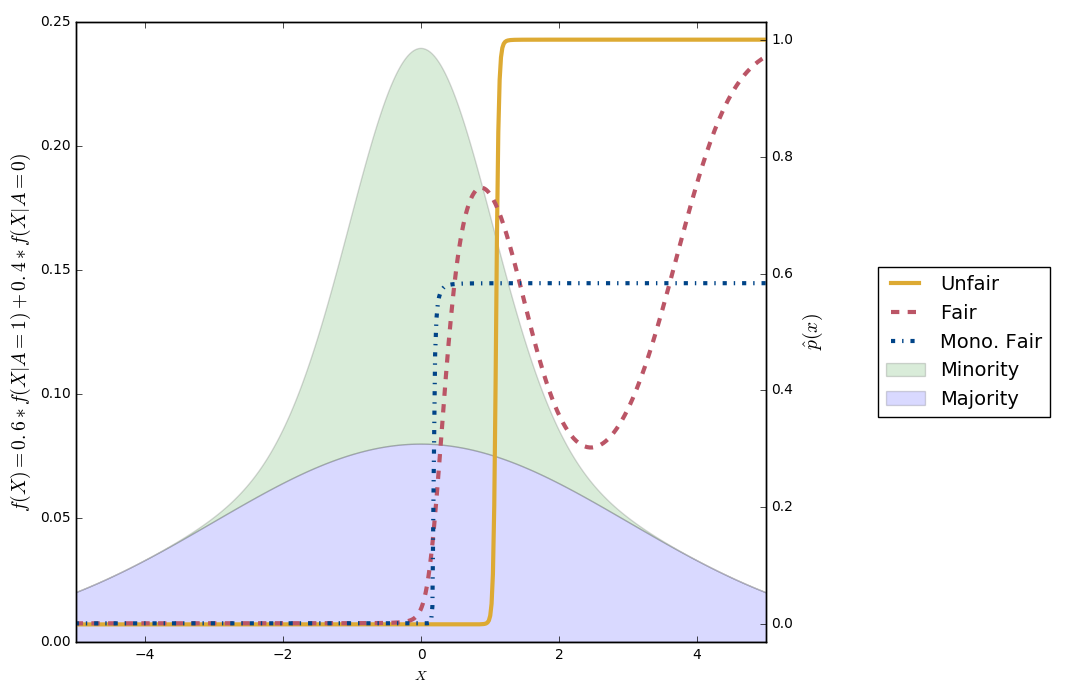}
        \caption{The distribution of $X$ for the minority class (light green, $X|A=0 \sim N(0, 1)$) differs from that of the majority class (light blue, $X|A=1 \sim N(0, 3)$). We have $P(A=1) = 0.6$. For both classes, $Y = X + \epsilon$, $\epsilon ~ N(\mu=0, \sigma=0.1)$  -- i.e. the chance of success increases with $X$. "Unfair" (yellow solid line) is an unconstrained neural network soft classifier which maximizes expected outcome score of positive predictions subject to a constraint on expected number of positive predictions. "Fair" (red dashed line) adds the restriction that we must have equal expected probability of positive prediction for both classes.  
        "Mono. Fair" (dark blue dash-dot line) adds the further constraint that the prediction function must be monotonic.)}
        \label{fig:nonmono_problem}
    \end{figure*}
    
    The synthetic example in Figure~\ref{fig:nonmono_problem} demonstrates such a situation. We consider the setting where we wish to create a soft classifier, $\hat{p}_i = f(X_i)$ which maximizes the average score of positive predictions $\sum_i \hat{p}_i Y_i$ with a constraint on the expected number of positive classifications $\sum_i \hat{p}_i$---this might correspond to admitting a fixed number of students based on their predicted future performance. The true relationship is that $Y \sim  N(X, \epsilon)$. Our classes are imbalanced and have different distributions, as shown in Figure~\ref{fig:nonmono_problem}. An "unfair" classifier that does not aim to achieve demographic fairness, learns a hard threshold at $X=1$ but leads to 2.58 times higher odds of acceptance for the majority class vs.\ the minority class.
        
    We can achieve a more fair result by adding a penalty that encourages demographic parity\cite{HarPriSre2016}, which requires that the probability of a favorable outcome be independent of class, i.e. 
     $\frac{\sum_{i: A_i = 0} \hat{p}_i}{\sum_{i: A_i = 0} 1} = \frac{\sum_{i: A_i = 1} \hat{p}_i}{\sum_{i: A_i = 1} 1}$.  Adding such a penalty reduces the odds ratio from 2.58 to 1.13.
    However, in order to maximize demographic parity, the fair classifier ends up learning a non-monotone function. All those with $X \in (0.9, 4.0)$ receive predictions lower than those with $X = 0.9$ regardless of protected attribute. Clearly, this would lead individuals in the region to resent individuals with lower attribute values: individuals in this range would have a better chance of a positive outcome if they had a ``worse'' value of $X$.
        
    By contrast, a monotonically fair classifier ("Mono. Fair") learns a function that avoids the score resentment present in the "Fair" classifier, while achieving similar demographic parity (odds ratio 1.11). No individual can claim that another individual with a lower non-protected attribute value received a higher probability of acceptance.  This is achieved by reducing the certainty of acceptance from those with the highest attribute values, which are increasingly majority-dominated, and reducing the threshold attribute value required to have any chance of acceptance. 
        
   If we add in the requirement that our score function is Lipschitz-continuous, we can see monotonic fairness as an extension of individual fairness. Where $X^\circ \neq \emptyset$ and we have non-protected attributes that do not require monotonicity, incorporating a Lipschitz requirement avoids seemingly arbitrary discontinuities across $X^\circ$. Where $X^\circ = \emptyset$ and where we require monotonicity along all dimensions of $X$, the Lipschitz requirement is likely to be less important, since any discontinuities will favor higher-valued attributes. Further, enforced monotonicity will likely lead to smoother functions with fewer discontinuities than non-monotone solutions.
    
\section{Learning monotonic fair scores using neural networks}
    
    As described above, any score function whose value does not depend on the protected attribute, and that is monotonically non-decreasing with each dimension of $X^+$, will have zero individual resentment under the conditions discussed in Section~\ref{sec:defs}.\footnote{In this section, we only consider the monotonically non-decreasing case; the monotonically non-increasing case can be considered analogously.} A number of algorithms have been proposed to learn monotone functions; \cite{cano2019monotonic} offers a detailed review. We choose to use feedforward neural networks, since they are flexible and easily adapted to a specific problem.
    
    We restrict our analysis to situations where value comparisons are only made between individuals who differ in a single dimension of their non-protected attributes. In practice, this covers a large number of realistic use cases: it is easier for a practitioner to specify orderings in such settings. Ordinal categorical variables can be captured either by mapping the categories to integers, or by using dummy variables and setting the dummies for all categories worse than the actual category. We leave relaxation of these assumptions, and approaches for automatically learning orderings, to future work.  We also assume that ordering of attributes $X^k\in X^+$ correspond to some notion of ``value'', where we wish to impose the requirement that increasing $X^k$ does not decrease the chance of the more desirable outcome, provided the other attributes do not change, i.e. the relationship is monotonic. If necessary, the attributes may have been transformed by the practitioner to achieve this (e.g.\ mapping categories onto the reals).
    
    \begin{figure*}[t!]
        \centering
        \includegraphics[width=\textwidth]{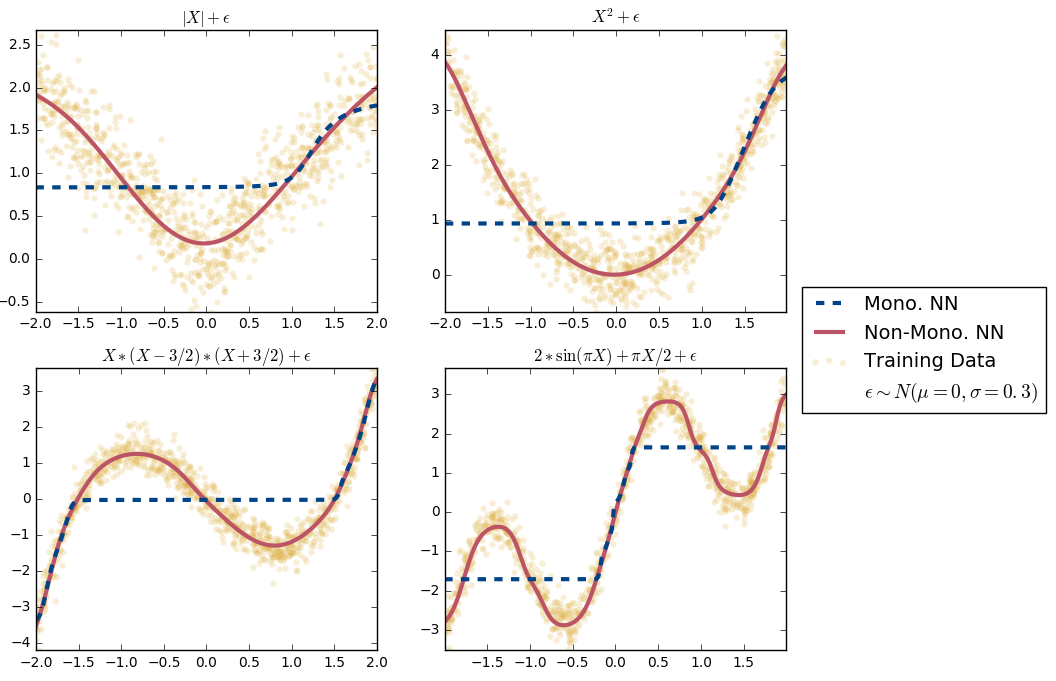}
        \caption{Training data (yellow circles, $n=1000$ for each), monotonic neural network (dashed blue line), and non-monotonic neural network with transformed weights after the first layer (solid red line) approximations for training data sampled from four example functions.}
        \label{fig:fourfunctions}
    \end{figure*}
    
     If we desire our function to be monotone non-decreasing with respect to every dimension of $X$, we can enforce this by ensuring all weights in the network are strictly positive, for example by applying some transformation $\tau:\mathbb{R}\rightarrow \mathbb{R}_+$ \cite{sill1998monotonic}. In the more general setting, where we wish to be monotone w.r.t.~$X^k\in X^+$ but do not require this for $X^k\in X^\circ$, partition the weights in our neural network into those that will be multiplied by (functions of) $X^+$, and those which will not. In a simple feedforward neural network setting, that means that in the first layer, weights corresponding to $X^k\in X^+$ are forced to be positive, while weights corresponding to $X^k\in X^\circ$ are not. In subsequent layers, all weights are required to be positive. Concretely, we apply the following transformations to the unconstrained weights $w_{\ell, k, i}$ of the neural network:
    
    \begin{equation} \tilde{w}_{\ell, k, i} = \left\{ \begin{array}{c l}
    \tau(w_{\ell, k, i})
                & ~\mbox{if}~ \ell > 1 \mbox{ or } X^k \in X^+ \\ 
    w_{\ell, k, i} 
                & ~\mbox{if}~ \ell = 1 \mbox{ and } X^k \in X^\circ \\ 
        \end{array} \right. \label{eqn:transformation} \end{equation}
    \begin{equation}h_{\ell, k} = \sigma\left(\sum_i \tilde{w}_{\ell, k, i} h_{\ell-1, i} + b_{\ell, k}\right)\, .\label{eqn:monotonic_nn}\end{equation}
    
    The output is clearly a monotone non-decreasing function\footnote{We assume the use of an activation function which is also monotone non-decreasing, which is common (e.g. ELU, ReLU, leaky ReLU, tanh, sigmoid) but not universal.} of each $X^k\in X^+$, since all weights in the path of such $X^k$ are positive. Leaving $w_{1, k, i}$ unconstrained for $X^k\in X^\circ$ allows for the function to be non-monotonic w.r.t.\ those $X^k$.

    In our experiments, we use an offset form of the exponential linear unit~\cite{clevert2015elu} transformation, 
    \begin{equation}\tau(x) = \left\{\begin{array}{c l} 
        x       & ~\mbox{if}~ x > 1 \\ 
        e^{x-1} & ~\mbox{if}~ x \le 1 \\ 
    \end{array}\right.,\label{eqn:tau}\end{equation}
    in Equation~\ref{eqn:monotonic_nn} to transform the appropriate weights to be positive. Note that any continuously differentiable function with strictly positive range could be substituted; we selected the offset exponential linear unit based on experimental performance. We explore other choices in the supplement.  

    Figure~\ref{fig:fourfunctions} explores the effect of the transformations $\tau$. We show the outputs of two neural networks: One where all weights are transformed according to Equation~\ref{eqn:tau} (Mono. NN), and one where the first layer is untransformed but subsequent layers are (Non-Mono. NN). The first network demonstrates that this architecture is able to learn monotonic functions even when the true function is non-monotone. The second network demonstrates that, provided the first layer is not transformed, the transformation of weights in subsequent layers does not interfere with fitting arbitrary functions with the usual precision (and drawbacks) of feedforward neural networks. Since we can arbitrarily transform the edge weights between a subset of the inputs and the first layer, we can also fit higher-dimensional functions which are monotonic only on a subset of the inputs.  See the supplement for two-dimensional examples.
  
    Neural networks have been used to learn fair classifiers in a number of contexts \citep{LouSweLiWelZem2016,BeuCheZhaChi2017,MadCrePitZem2018,xu2018fairgan}. Dwork et al. \cite{dwork2012fairness} originally posited individual affirmative action within a framework of Lipschitz smoothness. In many commonly used architectures (including the ones used in this paper), neural networks describe Lipschitz-continuous functions, although the Lipschitz constant may be large \citep{Szegedy:2014,gouk2018regularisation,Balan:Singh:Zou:2018}. One could also enforce greater smoothness by Lipschitz continuity-aware regularization \citep{gouk2018regularisation}. We choose not to do so in our experiments, relying on the monotonicity constraints to add additional regularization, to ensure that any jumps (w.r.t.~$X_k\in X^+$) are individually fair, and to enforce that the effective decision rule does not create the potential for resentment.

    In addition to monotonic fairness, we also want to ensure our algorithm has desirable group-level fairness properties. To do so, we train our monotonic neural network using backpropagation to minimize a compound loss
    $$\mathcal{L}(\theta) = \lambda_P \mathcal{L}_P(\theta) + \lambda_F \mathcal{L}_F(\theta)$$
    evaluated on a minibatch, where $\mathcal{L}_P$ is a prediction loss, $\mathcal{L}_F$ is a fairness loss, and $\lambda_P, \lambda_F \ge 0$ are weights governing the relative importance assigned to each loss. 
    
    The fairness loss, possibly derived from a constraint, encourages a desired form of fairness, and is calculated across the entire minibatch. A variety of differentiable losses have been developed that could be deployed here \cite{KamAkaSak2011,KamAkHidSak2012,berk2017convex,zafar2017parity,zafar2017aistats,AgaBeyDudLanWal2018}. In our experiments, we use the demographic loss proposed by \cite{ZemQiSwePitDwo2013}, $|\bar{y}_0 - \bar{y}_1|$, i.e.\ the absolute difference in mean prediction between majority and minority classes.  
    
    The prediction loss is some loss that penalizes predictions that are far from ground truth, for example cross-entropy or MSE. This loss is typically evaluated individually for each data point, and then summed over the minibatch.

\section{Experiments}
        We evaluated\footnote{A python implementation is available at\\ \href{https://github.com/throwaway20190523/MonotonicFairness}{https://github.com/throwaway20190523/MonotonicFairness}} our method on three real-world examples of increasing complexity: law school admissions, COMPAS scoring of recidivism risk in bail decisions, and German credit assessment in granting loans.  In each case, both our protected variable $A$ and our target $Y$ are binary. We specify our compound loss as a convex combination of cross-entropy and equality of outcome, following the example of~\cite{ZemQiSwePitDwo2013}, though other measures are interchangeable if they are differentiable. Concretely, for a minibatch  $\mathcal{M} = (X_i, Y_i, A_i)_{i=1}^M$, we have:
        {\footnotesize
            \begin{equation}
            \begin{array}{l}
            \mathcal{L}(\theta;\alpha, \mathcal{M}) =
            \alpha \underbrace{
                    \left| 
                        \frac{\sum\limits_{i: A_i = 1} \hat{p}(X_i; \theta)}{\sum\limits_{i: A_i = 1} 1}
                        -
                        \frac{\sum\limits_{i: A_i = 0} \hat{p}(X_i; \theta)}{\sum\limits_{i: A_i = 0} 1}
                    \right|
            }_{\mathcal{L}_F} + \\
            (1 - \alpha) \underbrace{\frac{1}{M}
                \sum_{i=1}^M -\left(Y_i \log\left(\hat{p}(X_i; \theta)\right) + (1 - Y_i) \log\left(1 - \hat{p}(X_i; \theta) \right)\right)
            }_{\mathcal{L}_P} \nonumber 
            \end{array} \label{eqn:experiment_loss}
            \end{equation}
        }
        where $\hat{p}(X_i; \theta)$ is the output of our neural network, and $\alpha\in (0,1)$ controls the balance between fairness and prediction.

        We compare against both a neural network with the same compound loss but no monotonicity constraints---which is representative of the set of individual-classifier methods described in Section~\ref{sec:bg}---and the Fair Representations method~\cite{ZemQiSwePitDwo2013}. The Fair Representations method establishes \textit{prototypes} for the data, each equipped with a location in data space and a mean outcome value, with actual data given a mixed membership vector to these prototypes based on a spherical Gaussian kernel. A penalty for demographic balance within each prototype's membership rate forces predictions to have demographic balance. This method achieves individual fairness since any two individuals with similar (unprotected) attributes will be given a similar outcome, and the mixed membership via kernels produces a Lipschitz-smooth outcome function.

\subsection{Datasets}
\begin{figure}[ht]
            \centering
            \includegraphics[width=.5\textwidth]{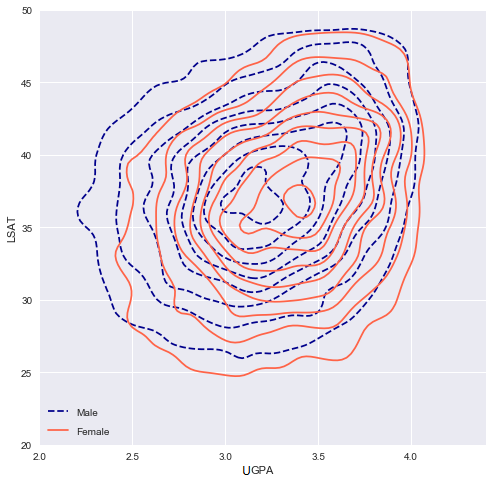}
            \caption{Distribution over UGPA and LSAT for male and female students. Female students tend to have higher GPA, but lower LSAT scores.}
            \label{fig:law_contour}
        \end{figure}
        
        \paragraph{Law school admissions data \citep{wightman1998lsac}:} This dataset  contains data from 9800 male and 7600 female law school students\footnote{The data has a pre-separated test set of 4,358 individuals; we additionally set aside 3,486 of the training examples as a validation set.} from 1991, with an outcome variable of normalized first year average (ZFYA) grades in law school and non-protected attributes of undergraduate grade point average (UGPA) and LSAT score (LSAT).\footnote{The LSAT exam has undergone extensive change since this data was collected in 1991.  Our analysis is motivated by the real-world dataset, but our conclusions are not necessarily applicable to the current exam. In addition, the dataset is limited to individuals admitted to law school and is not a representative sample of all test takers (many of whom would not have an observed outcome).}  We use gender as our protected attribute, and binarize the outcome by setting $Y=1$ whenever $\mbox{ZFYA} \ge 0.09$, its median value. One result with an apparently erroneous UGPA of 0.0 was removed before analysis. Figure~\ref{fig:law_contour} shows contour plots of the per-gender non-protected attribute distributions, generated by adding uniform noise to counter the discretization of the data then using kernel density estimation. We see that female students tend to have higher GPA, but lower LSAT scores, than the male students (see Figure~\ref{fig:law_contour}).
        
        \paragraph{COMPAS data \cite{larson2016we}:} Released in 2016 following a public interest investigation into machine learning methods in criminal justice, the COMPAS dataset (named for the proprietary system which generated it) contains the risk factors, demographic information, and two-year recidivism information for over 7,000 individuals arrested in southern Florida in 2013 and 2014.  We reduced this to a two-class problem by restricting our analysis to the 6,150  ``African American" and ``Caucasian" examples in the dataset,\footnote{We set aside 1,235 as a test set, and 658 as a validation set for the neural network models.} and attempt to predict the two-year recidivism risk of the accused based on their age (non-monotonic) and number of prior adult convictions, juvenile felony, misdemeanor, and other convictions (all monotonically non-decreasing).
        
        \paragraph{German credit data \cite{ucigerman}:} Covers 1,000 credit applicants in Germany,\footnote{We randomly select 20\% (200) to use as a test set, and 20\% of the training set (160) are set aside by the neural network models for validation data.} including their employment, financial, and residency information, as well as the type of loan they requested and whether they repaid it.  We treat age (already binarized by the data source) as the protected attribute. There are 58 attributes in the dataset, of which we converted 7 into monotonic numeric variables: (monotonic non-decreasing) current checking account balance, credit history, employment tenure, and savings balance, and (monotonic non-increasing) investment as income percentage, length of loan in months, and credit amount. In the case of monotone non-increasing inputs, the corresponding weights in the first layer are transformed to be negative, rather than positive.   These were done intuitively, based on the idea that no one should be penalized for having more money in reserve, more stable employment, or better credit history, and no one should be rewarded for increasing the borrowed amount or requesting more months to pay it back, holding all other things constant.

        \subsection{Models}
        For each dataset, we trained three models:
        \begin{itemize}
            \item FNN: A non-monotonic, feedforward Fair Neural Network with 4 hidden layers of 10 nodes and \textit{tanh} activation functions\footnote{For monotonic networks, an activation function with bounded range is useful in order to allow the function to be non-convex; see supplemental materials.} using an ADAM optimizer.
            \item FMNN: A Fair Monotonic Neural Network otherwise identical but with monotonically-transformed weights where appropriate.
            \item FR: Fair Representations \cite{ZemQiSwePitDwo2013} with 10 prototypes. 
        \end{itemize}
        
        For each model, we trained 100 versions of the model with $\alpha$ randomly sampled according to a  $\mbox{Beta}(0.5, 0.5)$ distribution.  This distribution allowed us to heavily sample near the bounds to accommodate imbalanced losses.  For the FR model, we also randomly sampled a value for their coverage penalty $A_x$ from a log-uniform distribution between $10^{-2}$ and $10^2$ (and setting $L_Y = 1 - \alpha$ and $L_Z = \alpha$).  All datasets were scaled to have marginal variance of 1 for all input dimensions, as unequal scales can affect coverage statistics.  For the neural network models, minibatching (size 256 for COMPAS and Law School, 128 for German) and stepwise scoring on a 20\% validation subset (taken from the training data) were used to prevent overfitting.
   
    \subsection{Results}
        \begin{figure*}[h!]
            \centering
            \includegraphics[width=\textwidth]{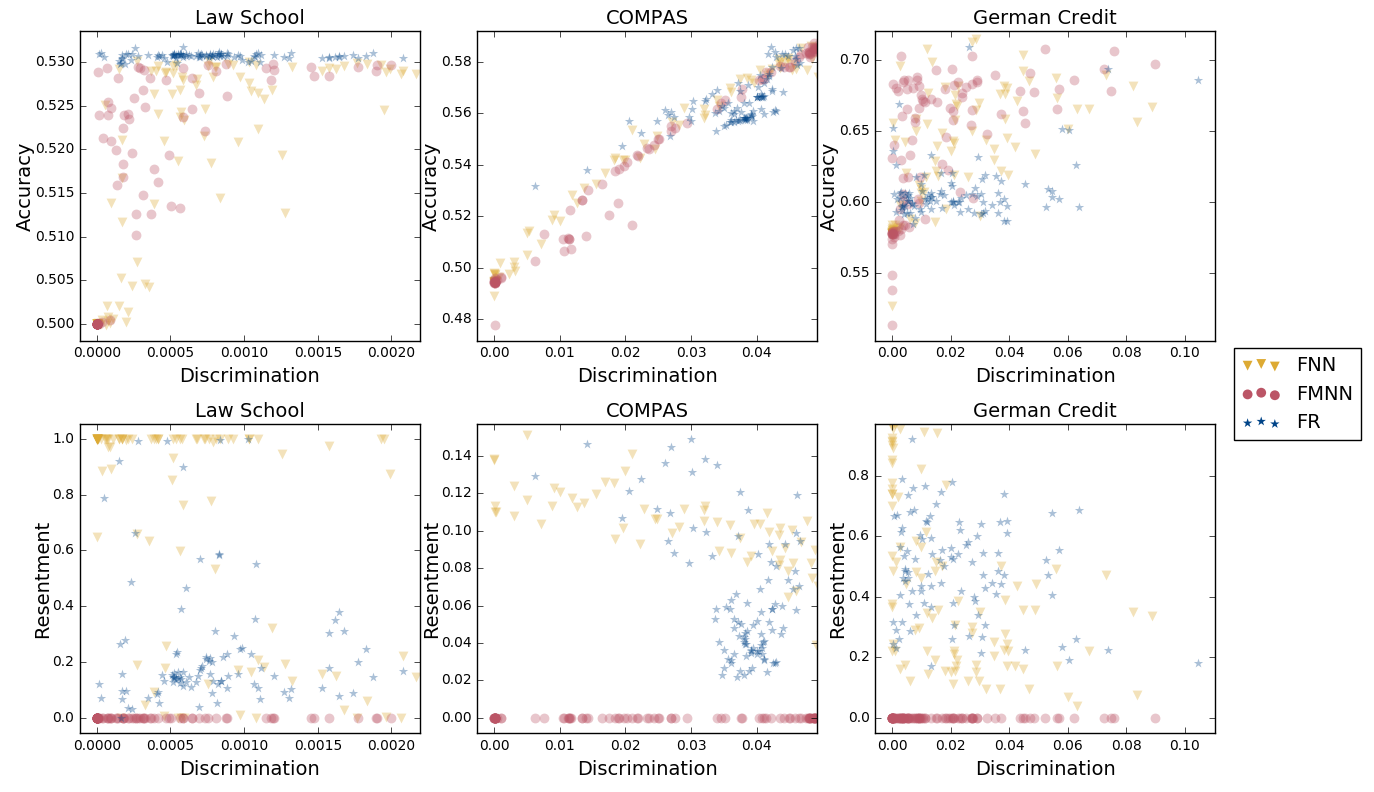}
            \caption{Accuracy vs Discrimination (top row) and Discrimination vs. Resentment (bottom row) across models and datasets.  Yellow triangles are FNN, red circles are FMNN, blue stars are FR.}
        \label{fig:tradeoffs}
        \end{figure*}
        
        \begin{figure*}[h!]
          \centering
          \includegraphics[width=\linewidth,trim={0 3cm 0 3cm},clip]{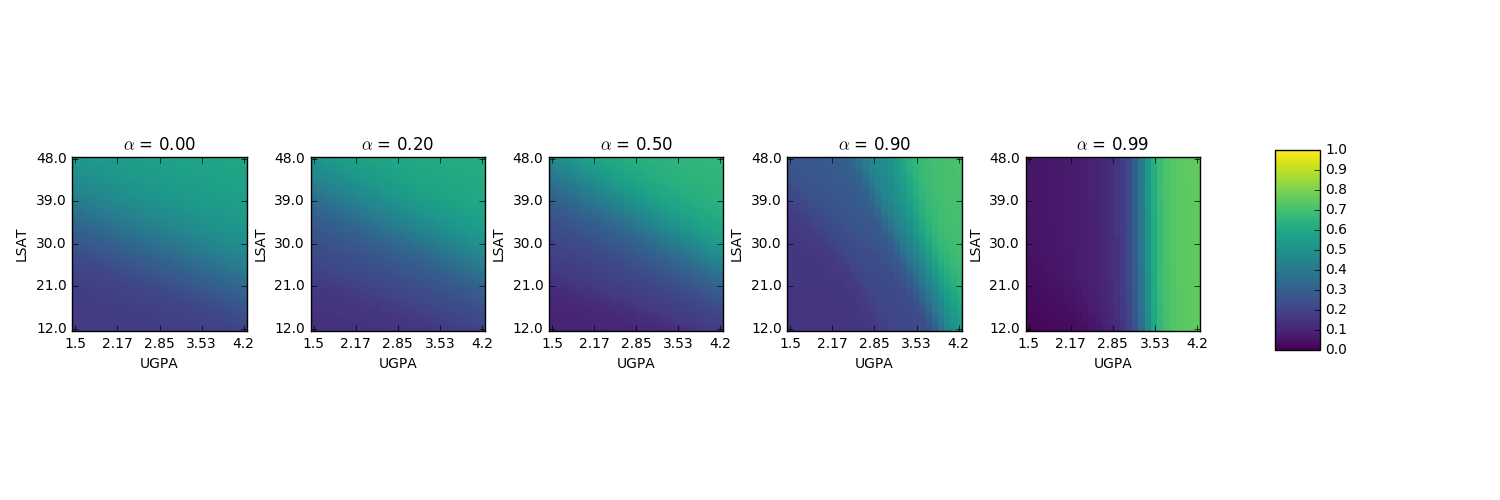}
          \includegraphics[width=\linewidth,trim={0 3cm 0 3cm},clip]{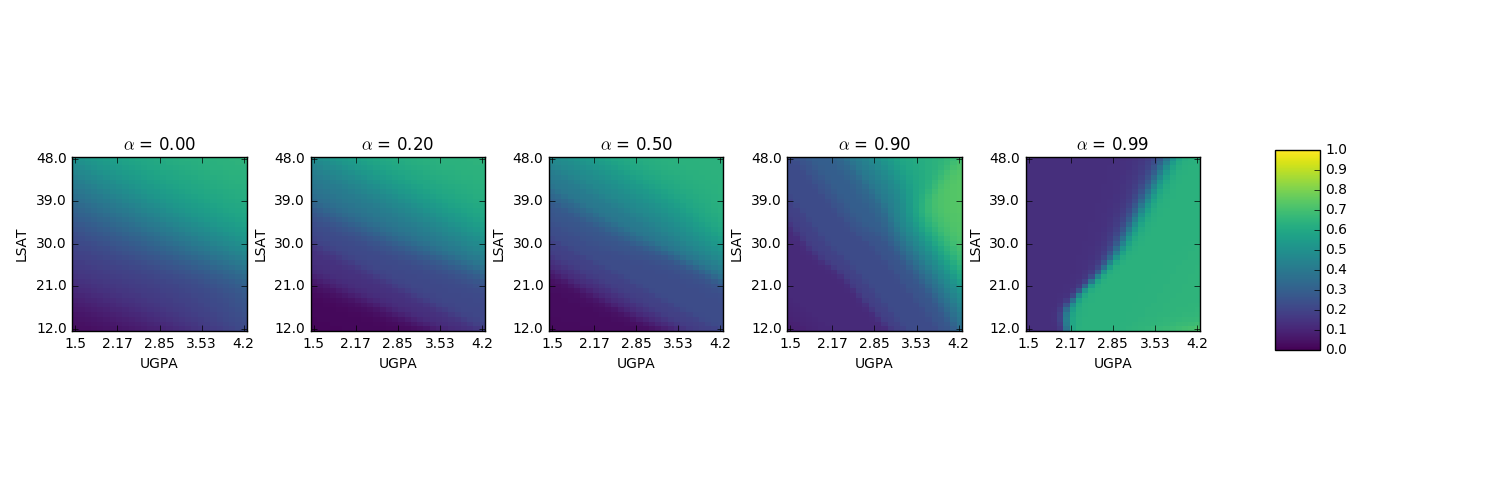}
            \caption{Plots of fitted solution for law school admissions data across range of $\alpha$ (fairness) levels, with unfairest left and fairest right.  Top row: Monotonically fair classifier. Bottom row: Classifier with no monotonicity constraint. Lighter color indicates higher value.}
            \label{fig:fit_plots}
        \end{figure*}

        In Figure~\ref{fig:tradeoffs} we see the usual accuracy-discrimination trade-off in the upper row of plots.  Accuracy and discrimination are defined as in \cite{ZemQiSwePitDwo2013}:
        \begin{itemize}
            \item Discrimination: $ \left|
                \frac{\sum_{n:s_n=1} \hat{y}_n}{\sum_{n:s_n=1} 1} -
                \frac{\sum_{n:s_n=0} \hat{y}_n}{\sum_{n:s_n=0} 1}
                \right| $
                \vspace{0.1in}
                
            \item Accuracy: $1 - \frac1N \sum_{n=1}^N |y_n - \hat{y}_n|$
        \end{itemize}
        
        In most cases, we see that the monotonic neural network is of similar or slightly lower accuracy than the non-monotonic neural network or the Fair Representations approach for a given level of discrimination.   This is unsurprising, since the non-monotonic methods are free to learn an unconstrained function.  We would only expect the monotonic method to yield better predictions if the underlying data has a strictly monotonic generating function.  However, we see that the loss in accuracy is generally small and likely tolerable across all three example datasets.
        
        In the bottom row of plots in Figure~\ref{fig:tradeoffs}, we see a different trade-off: the cost in individual resentment for improving group fairness.  Here, resentment is measured as the proportion of individuals in the test set who experience individual resentment, as defined in Section~\ref{sec:defs}. Specifically,
        \begin{itemize}
            \item Resentment: $ \frac1N \sum_{i=1}^N \max\limits_{j \in \mathcal{N}_i} \left(1_{\hat{y}_i < \hat{y}_j}\right)$
        \end{itemize}
        where $\mathcal{N}_i$ is the set of $j \neq i \in \{1 \ldots N\}$ where $X_i$ is ``better" than $X_j$ or $X_i = X_j$ and $A_i\neq A_j$.
        In practice, since none of the methods use the protected attribute as an input, this is equivalent to the number of individuals who experience non-protected attribute (score) resentment, i.e.\ they had a higher attribute in a monotonically non-decreasing dimension (or a lower one in a non-increasing dimension) than a hypothetical individual with a more favorable prediction (and identical non-protected attributes). 
        
        Due to the high dimensionality of some of the datasets, we restricted our consideration of resentment to individuals who feel resentment towards a peer in the test set, rather than resentment towards a hypothetical individual with worse scores.   Note that, as the dimension of the attribute space increases, the sample estimate will underestimate resentment, due to a decreasing number of individuals with comparable attributes.  For example, in the law school admissions setting, it is easy for an individual to find peers with lower UGPA but the same LSAT scores; conversely, for the German credit data, a comparable individual must match on 51 attributes.  However, the resentment of the monotonic neural network will always be zero by design.  
        
        \begin{figure*}[t!]
        \centering
        \includegraphics[width=\textwidth]{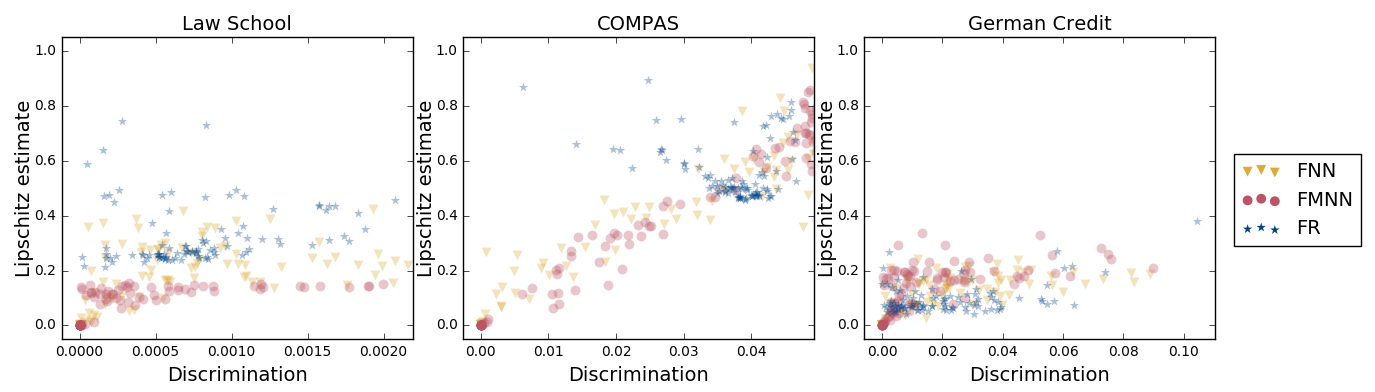}
        \caption{Lipschitz constant estimate vs.~discrimination across models and datasets.  Yellow triangles are FNN, red circles are FMNN, blue stars are FR.}
    \label{fig:lipschitz}
    \end{figure*}
    
        Let us explore the Law school dataset in more detail. Figure~\ref{fig:law_contour} shows the comparative distributions of males and females w.r.t.\ GPA and LSAT score.  Note that the female distribution is shifted towards higher UGPA and lower LSAT score than the male distribution. In Figure~\ref{fig:fit_plots}, we see the admissions probabilities produced by the monotonic and non-monotonic neural networks. When $\alpha$ is high, we see that individuals would often do well to \textit{lower} their reported LSAT score in order to increase their probability of admission.  This is an artifact of the disproportionate number of women with high UGPA and low LSAT scores, resulting in a ``fair" classifier which favors lower LSAT scores for individuals high UGPA, similar to the example in Figure~\ref{fig:nonmono_problem}.  Even though there is no resentment across protected variable groups, there clearly would be resentment by those who are less likely to receive a favorable outcome due to a counter-intuitive admissions policy designed to produce demographic balance. 

        \subsubsection{Lipschitz constant}
        
         Although our method is not primarily intended to produce a smoother function, i.e. one with a lower Lipschitz constant, it is a desirable property for individually-fair functions.  Zhang et al. \cite{zhang2019recurjac} provide a discussion of the advantages and disadvantages of several types of empirical estimators of the Lipschitz constant for a neural network.
        
        We adopt a sample-based estimator similar to that of \cite{wood1996estimation}, which uses a pairwise evaluation of the constant,\footnote{The method proposed by \cite{wood1996estimation} further fits estimates a parametric distribution of the values to find an estimate of the maximum, but that method requires a random sample  of points which is infeasible here.  We instead use the maximum of empirical distribution, which is biased downwards but adequate for comparison purposes.} i.e. $$ \hat{L} = \max\limits_{i, j} \left( \left| \frac{\hat{Y}_i - \hat{Y}_j}{ d(X_i, X_j) } \right|\right)$$  
        
        We calculate our Lipschitz constant with respect to a standardized Euclidean distance, 
        
        $$d(X_i, X_j) = \sqrt{\sum\limits_k \left( \frac{X_i^k - X_j^k}{\hat{s}_k}\right)^2}$$
        
        where $\hat{s}_k$ is the sample standard deviation of $X^k$. We standardize in this manner so smoothness is comparable across dimensions.   As discussed in \cite{zhang2019recurjac}, this sample estimate is a lower bound of the true constant, but we feel it is adequate for model comparison.
        
        In Figure~\ref{fig:lipschitz}, we see that the monotonic neural network tends to produce smoother solutions for a given value of discrimination than other methods in more inherently-monotonic settings like the Law School dataset than in less inherently-monotonic settings like COMPAS or German Credit.  This is unsurprising, since the monotonicity constraint acts as a regularizer, preventing overfitting to spurious non-monotonic trends in sampled data.

    \section{Discussion}
    Individually fair classifiers can exhibit unfair behavior on a population level, and can lead to the undesirable situation where an individual who performed worse on a given metric would have had a better outcome, leading to resentment. We show that a definition of individual fairness that incorporates monotonicity can avoid the latter situation, and can be combined with measures of demographic fairness to yield classifiers that trade off predictive power with demographic fairness.
    
        Several recent works suggest important future directions.
        
        \paragraph{Estimation of monotonic relationships:} A critical requirement of individual fairness as originally proposed \cite{dwork2012fairness} is a distance metric over $\mathcal{X}$ to determine the degree of similarity between individuals.  Our work sidesteps the problem by relaxing the requirement from a distance metric to a concept of ordering.  Recent concurrent works \cite{jung2019eliciting,ilvento2019metric} have explored the concept of estimating a distance metric by polling fair experts on what constitutes similarity.  We can similarly imagine extending the current work by polling fair experts instead on which individuals should receive higher outcomes than others, and enforcing coherence between the trained prediction function and the poll results on orderings.  This would allow one to relax the requirement of explicitly monotonic dimensions in the input data.
        
        \paragraph{Post hoc adjustment for monotonicity:} Recent works, e.g. \cite{lohia2019bias}, have attempted to use post hoc adjustments and model pooling prevent biases in machine learning.  These methods approach machine learning methods as black box function estimators, and instead of modifying the input data or function space of the models, use post hoc adjustment of the trained models' predictions in order to create fairness.  It is reasonable to consider whether we can extend this general applicability to the current approach; if we have a classifier which satisfies other concepts of fairness and accuracy, we may be able to manipulate its outputs to induce monotonicity on their outputs without interfering in the ``black box."

\clearpage
\bibliography{fairness}
\bibliographystyle{plain}

\clearpage

\section*{Supplement}
    In this supplement, we provide justification for our design choices for the neural network architecture, and demonstrate that such an architecture is able to capture monotonic functions, and impose monotonicity even when the true generating function is non-monotone.
    
    \subsection*{Design choices}
    Below, we discuss several design choices, and their effect on the resulting functions.
    
\begin{figure*}
                \centering
                \includegraphics[width=\textwidth]{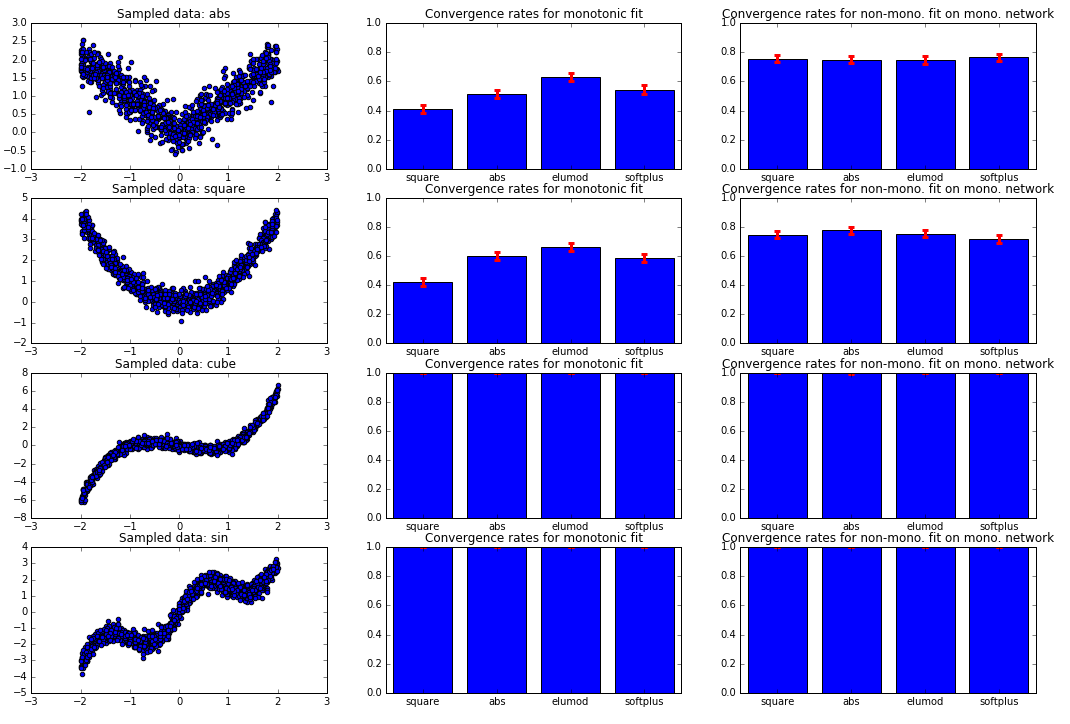}
                \caption{Convergence rates for various functions used to enforce positive weights. The vertical exist for the middle and right columns is the proportion of random initialization which converge to a non-deviant ($\hat{y} = \bar{y}$) solution.}
                \label{fig:activations}
            \end{figure*}

        \paragraph{Transformation Matters:} The choice of transformation function in Equation 4 can have a significant effect on the probability of successful convergence of monotonic neural networks.  We show in Figure~\ref{fig:activations} that the choice of transformation can have different effects based on the nature of the underlying function, and affects both monotonic and non-monotonic fitting. We consider four non-linearities:
        \begin{itemize}
            \item Square: $\tau(x) = x^2$.
            \item Abs: $\tau(x) = |x|$.
            \item Offset exponential linear unit (elumod): \\$ \tau(x) = \left\{\begin{array}{c l} 
        x       & ~\mbox{if}~ x > 1 \\ 
        e^{x-1} & ~\mbox{if}~ x \le 1 \\ 
    \end{array}\right.$
            \item Softplus: $\tau(x) = \log(1+e^x)$
        \end{itemize}
        We choose to use an offset exponential linear unit in our experiments, since it achieved optimal or near-optimal convergence in these comparisons.
        
        \paragraph{Activation Matters:} Additional caution is needed in selecting an activation function for a monotonic neural network.  If, for instance, a convex activation function is used (e.g. \textit{elu} or \textit{relu}), subsequent layers can only compound this convexity, and the resulting function can only be convex.  It is easy to see this by considering the compounding of the first and second derivative across the layers.  This may be a desirable feature in some settings, but generally prohibits it from approximating \textit{any} monotonic function.  As such, bounded (but monotonic) activation functions like \textit{logistic} or \textit{tanh} are advisable for general purposes.
        
    \subsection*{Ability to Capture Mixed Monotonicity}
        We wish to emphasize that the network architecture described in this paper can simultaneously handle monotonic and non-monotonic relationships between the inputs and output. If we begin with the assumption that a network constrained to positive weights will produce a monotonically increasing function $f(x)$, we can briefly intuit the ability to fit a monotonically decreasing function by considering that $f(-x)$ would produce an identical function $f(x)$ but with reversed domain and therefore would be monotonically decreasing. Equivalently, we can enforce negativity on the weights in the network on edges leading out from any $x$ with respect to which $f(x)$ is monotonically decreasing, i.e. set $\tilde{w} < 0$ in the connection between $x$ and the first hidden layer (but keeping all weights in subsequent layers positive to maintain direction).  
        
        Further, if we accept that we can fit monotonically increasing and decreasing functions by constraining the weights, then consider what would happen if we fit $f(x, x)$, i.e. fed the same input twice, but constrained the first to be increasing and the second to be decreasing.  By the argument of decomposing functions into positive and negative parts (or, here, decomposing the first derivative into positive and negative parts), we can construct a monotonic function from its increasing and decreasing parts.  Further, each node in the first hidden layer would compute as $\sigma(\tilde{w}_{+} x + \tilde{w}_{-} x + c)$, which could be simplified as $\sigma(w x + c)$ where $w$ is unconstrained. 
        
        \begin{figure*}
                \centering
                \includegraphics[width=.8\textwidth]{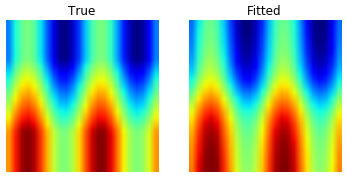}
                \caption{Demonstration of our network architecture's ability to fit a function which is monotonic in one dimension and non-monotonic in another. }
                \label{fig:demo1}
            \end{figure*}
            
        To demonstrate the result empirically, we show in Figure~\ref{fig:demo1} a two-dimensional experiment in which the true underlying function is non-monotonic w.r.t to $x_1$ but strictly monotonically increasing w.r.t. $x_2$.  Specifically,
        
            $$ f(x_1, x_2) = \mbox{sin}(\pi x_1) + \mbox{max}(-1, \mbox{min}(1, x_2))  $$
            
        The estimated function shown is fit on a sample of 1,000 samples from the function and set to be non-monotonic w.r.t. $x_1$ and monotonic w.r.t $x_2$ and is able to recover the true function with reasonable precision.
        
\begin{figure*}
                \centering
                \includegraphics[width=.8\textwidth]{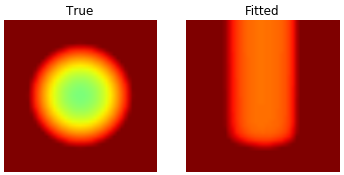}
                \caption{Demonstration of our network architecture's ability to created a function which is monotonic in one dimension and non-monotonic in another, even when the data does not meet those qualifications.}
                \label{fig:demo2}
            \end{figure*}

        Similarly, we show in Figure~\ref{fig:demo2} that a mixed-monotonicity function can be fit even if the underlying function is severely non-monotonic (with the expected error in fit).  Here, $f(x_1, x_2) = x_0^2 + x_1^2$, and we again fit on a sample of 1,000 samples from the function and set to be non-monotonic w.r.t. $x_1$ and monotonic w.r.t $x_2$.  As expected, it finds a function which is optimal subject to the (incorrect) constraints.
        
\end{document}